\documentclass[12pt,titlepage]{article}

\usepackage{indentfirst}
\usepackage{amssymb}
\usepackage{amsmath}
\usepackage{algpseudocode}
\usepackage{graphicx}
\usepackage{caption}
\usepackage{subcaption}
\usepackage{setspace}
\usepackage{booktabs}
\usepackage{pdfpages}
\usepackage{geometry}

\begin{document}

\renewcommand{\thefootnote}{\alph{footnote}}

\pagenumbering{gobble}

\begin{center}

\vspace*{\fill}

{\LARGE The Effect of Heteroscedasticity on Regression Trees}
\\[.5 in]

 Will Ruth\footnotemark[1] (Corresponding Author) and Thomas Loughin\footnotemark[2]

\vspace*{\fill}

\footnotetext[1]{Simon Fraser University. 8888 University Dr.~, Burnaby, B.C., Canada. V5A 1S6. wruth@sfu.ca.}

\footnotetext[2]{Simon Fraser University. 8888 University Dr.~, Burnaby, B.C., Canada. V5A 1S6. tloughin@sfu.ca}

\end{center}

\begin{abstract}
Regression trees are becoming increasingly popular as omnibus predicting tools and as the basis of numerous modern statistical learning ensembles. Part of their popularity is their ability to create a regression prediction without ever specifying a structure for the mean model. However, the method implicitly assumes homogeneous variance across the entire explanatory-variable space. It is unknown how the algorithm behaves when faced with heteroscedastic data. In this study, we assess the performance of the most popular regression-tree algorithm in a single-variable setting under a very simple step-function model for heteroscedasticity. We use simulation to show that the locations of splits, and hence the ability to accurately predict means, are both adversely influenced by the change in variance. We identify the pruning algorithm as the main concern, although the effects on the splitting algorithm may be meaningful in some applications.   \\[.1 in]

\end{abstract}


\section{Introduction}
\label{int}

\pagenumbering{arabic}

The regression problem is one that arises regularly in a wide range of areas. The classical solution to regression problems is to use a linear model with a least squares solution for the parameters \cite{kut04}. This technique is simple, powerful, and provides a reasonable approximate answer to many regression problems. However, it suffers from a number of shortcomings. In particular, it requires that the relationship between the response and any predictor variables be specified before any analysis can be conducted. This can hinder its use as an exploratory tool, particularly when explanatory variables may relate to a response in complex ways that may depend on other variables.

There are numerous methods that make a linear regression more flexible.  These include splines \cite{fri91}, generalized linear models \cite{mcc89}, generalized additive models \cite{has09}, and regression trees \cite{bre84}.  Among these, regression trees are of particular interest because of their ability to adapt to complex interactions \cite{bre84} and for their use in predictive ensembles such as bootstrap aggregation (``bagging", \cite{bre96}) and random forests \cite{bre01}. Regression trees operate by recursively splitting the data into groups and then computing a response prediction in each group independently of the others. Methods for performing this splitting have been proposed by various authors (\cite{bre84}, \cite{cia91}, \cite{cha95}, \cite{ale96}, \cite{loh02}, \cite{su06}, \cite{hot06}). The best known among these is the recursive partitioning algorithm  of Breiman et al.~\cite{bre84}, which we denote by RPAB and describe in more detail in Section 2.  For many algorithms, in particular the RPAB, the `splitting' of the data into groups requires no knowledge about the structure of the relationship between the predictor and response variables.

While regression trees increase the flexibility of the modelling process compared to linear regression, each splitting algorithm has its own set of assumptions. One assumption that is implicit in the RPAB is that the response variance is constant throughout the entire dataset (see Section 2). This assumption is often violated in real datasets. There is often some systematic relationship between the mean and variance of the data \cite{car88}. Lower and upper bounds on measurements create situations where data from means near the bounds have less potential for variability than do data from more interior means. For example, when the size of an organism is being measured as it grows, it is almost always the case that the variability among immature specimens is smaller than the variability among fully grown ones.

The effect of non-constant variance (also called \textit{heteroscedasticity}) on least squares linear regression is well known (\cite{kut04}, \cite{fox08}). Parameter estimates obtained using ordinary least squares (OLS) are unbiased, but they have increased variance and thus are no longer optimal, even if all other model assumptions are met. Furthermore, fitted values in regions of low variability can be quite far from the true values, relative to the size of the local errors, because the homoscedastic OLS criterion does not distinguish between large errors made in region of high variability and those made in regions of low variability. As a result, confidence intervals for means and prediction intervals for observations over- or under-cover in regions of low or high variability, respectively. One solution to regression estimation under heteroscedasticity is to use a weighted least squares procedure, the implementation and consequences of which are well studied \cite{car88}.

The effect of heteroscedasticity on regression trees has not yet been studied. This is a serious gap in our understanding of trees, made more acute by the fact that  trees are the foundation of random forests and other ensembles, which have been shown to be powerful, all-purpose predictors that are both easy to use and accurate in a wide range of problems (\cite{bre01}, \cite{has09}, \cite{chi10}).  Because of the great potential and apparent widespread use of regression trees, it is imperative that the assumptions and operational characteristics of regression trees be understood.

The urgency of this research is amplified by the fact that researchers in many disciplines use trees daily in statistical analyses, often without knowing how they work. For example, one of us attended the annual meeting of the American Fisheries Society in September 2011. In a session on “Resource Management”, ostensibly one with no particular connection to Statistics, three of six unrelated speakers performed some of their statistical analyses using trees, admitting that the procedure was somewhat of a black box to them. The method is clearly in common use, and hence in common mis-use, by well-meaning scientists and other practitioners throughout the world. Providing users in all disciplines with better understanding of the strengths and weaknesses of tree-based regression will improve the quality and reporting of all research results that rely upon trees for their analytics.

The goal of this paper is to examine the effect of heteroscedasticity on the performance of regression trees formed using the RPAB. A simulation study is carried out on a number of very simple mean and variance structures that are chosen because it is easy to see what predictions should be made on them by a regression tree. First, the RPAB is described in more detail.  Subsequently, models used for the simulation are introduced and the results are presented.  A discussion follows to summarize the implications of the simulation findings.

\section{RPAB Summary}
\label{rs}

The RPAB and all information in this section are due to Breiman et al.~\cite{bre84}. The RPAB for regression can be applied to a dataset that contains any number of numeric or categorical predictor variables and one numeric response variable. This algorithm partitions the variable space into regions with similar response values and uses the sample mean within each partition as a piecewise constant predictor across the sample space. Partitioning is achieved by recursive application of an algorithm that splits data into two distinct subsets. Each split is chosen to give the largest reduction in an objective criterion among all possible splits. Once these splits have been chosen, superfluous splits are removed using a cross-validation procedure called ``pruning.''

For simplicity, we consider the case with one response and one numeric predictor variable, called $Y$ and $X$, respectively. The first goal of RPAB is to find a value $x$ that splits the data into an upper subset ($X>x$) and a lower subset ($X < x$). It does this by considering a representative for each possible split that can be made---i.e. all $x$ that lie halfway between consecutive distinct ordered values of $X$---and then choosing the corresponding split that results in the smallest sum of squared errors (SSE)  using each subset's sample mean as the predictor. The choice of SSE as a criterion is based on computational efficiency: Breiman et al.~\cite{bre84} present a way to examine all possible splits in $O(n^2)$ time. Note that minimizing SSE is equivalent to performing ordinary least squares estimation and thus implies the assumption that data are homoscedastic.

This procedure is applied recursively to each of the two subsets that are created by the chosen split until some stopping criterion is met. One stopping rule is to require that the best split decrease the SSE by at least a certain nominal amount (e.g., $1 \% $ of the original SSE). Alternatively, a minimum group size criterion can be set in which a split is not considered unless each of the resulting groups will have at least a certain number of elements.

This splitting procedure tends to over-fit data, so after it terminates, a ``pruning'' procedure is implemented to eliminate splits that do not improve prediction. This pruning consists of computing the cross-validated prediction error for a sequence of tree sizes that offer the best improvements in SSE-per-added-node in the original data. The model that has the lowest cross-validation error is then selected as the final choice. Alternatively, a ``1-SE" rule is often applied that selects the smallest tree with cross-validated prediction error within 1 standard error of the minimum cross-validated error.

If multiple explanatory variables are being considered, the RPAB examines all possible splits for each variable and chooses the one that provides the greatest reduction in SSE. Pruning is then done in the same way as above to obtain an optimal tree.

\section{Simulation Study}

\subsection{Overview}
\label{ove}

The goal of this simulation study is to examine how well regression trees built using the RPAB perform when faced with heteroscedastic data. To that end, we consider a problem with a single explanatory variable $X$ and response $Y$. We generate data from two simple models for the mean where the ideal behaviour of a regression tree is known. We then subject data from these structures to varying degrees of heteroscedasticity. This allows us to clearly measure the effect that heteroscedasticity has on the regression trees. The two mean models that we consider are constant and piecewise-constant in a monotone increasing pattern.  The latter of these is one that the recursive partitioning tree is designed to detect, while also resembling a linear regression. The variances also follow one of two structures, either homoscedastic or heteroscedastic. In the heteroscedastic structure, the half of the data with the largest values of $x$ has larger variance.

For each combination of mean structure and degree of heteroscedasticity, we measure the regression tree's ability to locate the correct splits and avoid making incorrect splits. We also measure how well it estimates the true mean values. The details of the simulation follow. We use a simulation study to examine the splitting and prediction behaviour of regression trees under heteroscedasticity because of the complex array of potential outcomes implied by the recursive nature of the algorithm.  While a more mathematical assessment of the first split is not difficult, carrying this approach further into the recursion becomes daunting when there are many possible ordered combinations of split locations that can be selected.

\subsection{Mean Structures}
\label{ms}

The general model we consider is one in which the predictor variable $X$ takes values $x=1,...,n$ and the response variable $Y$ is generated as $Y_x = \mu_x + \varepsilon_x$, where $\mu_x$ is the mean at $x$ and $\varepsilon_x$ is a normally distributed error term with mean 0 and variance $\sigma^2_x$. Furthermore, $\varepsilon_{x_1}$ and $\varepsilon_{x_2}$ are independent for $x_1 \neq x_2$. We choose $n=1000$ to offer the regression tree many opportunities to split.  In the case of the piecewise-constant mean, this allows us to clearly determine whether the algorithm is reacting appropriately to the changes in mean by splitting at or near the true jump points.

For the constant mean structure, $\mu_x$ is taken to be zero for convenience. Under this framework, a regression tree should make no splits. We use this to investigate whether the presence of heteroscedasticity impacts the location and frequency of spurious splits in the RPAB. In particular, does split frequency increase in areas of higher variance, or is a change of variance misinterpreted as a change of mean?

The second mean structure that is considered is a step function, $\mu_x = \lceil x/100 \rceil$ where $\lceil z \rceil$ is the ceiling function returning the smallest integer that is greater than or equal to $z$. This framework is chosen to approximate a linear trend, while still having obvious split locations for the regression tree. Under this framework, the recursive partitioning algorithm should ideally choose to split at the boundaries between each group. Any splits that do not occur at or near a boundary constitute errors under this framework, as do any boundary splits that are missed. We call each group of $x$ values with the same mean a ``segment.'' This means that we have 10 segments, each containing 100 observations.

\subsection{Variance Structures}
\label{vs}

We define $\sigma^2_x$ according to the following piecewise constant function
\begin{align*}
\sigma^2_x &= \begin{cases}
c_1^2 & \text{if } 1\leq x \leq 500\\
c_2^2 & \text{if } 501 \leq x \leq 1000
\end{cases}
\end{align*}
where $c_1,c_2=1,...,10$. For the constant mean model, it suffices to specify only the ratio $c_2/c_1$ because the SSE criterion is scale invariant \cite{bre84}. Thus, we typically fix $c_1=1$ and allow $c_2$ to take the values 1,...,10. Note that $c_1=c_2=1$ corresponds to homoscedasticity.

For the stepwise-increasing-mean model, both $c_1$ and $c_2/c_1$ matter because the ability to detect changes in mean is affected by scaling. As such, we consider all homoscedastic cases where $c_1=c_2=1,...,10$ as well as the heteroscedastic cases where $c_1=1$ and $c_2=2,...,10$. For reasons made clear later, we also consider homoscedastic cases where $c_1=c_2=\sqrt{(1+c^2)}/2$, $c=2,3,...,10$.

We use this simple variance structure so that the fundamental behaviour of the recursive partitioning algorithm can be explained under heteroscedasticity. The different magnitudes of $c_1$ and $c_2$ allow the effect of varying the signal-noise ratio in the data to be investigated. The homoscedastic variance structures are included as a control, against which results under heteroscedasticity can be compared.

We denote a given combination of mean and variance structure by its mean model (``$F$'' for flat or constant, ``$S$'' for stepwise-increasing), its variance model (``$H$'' for heteroscedastic, ``$C$'' for homoscedastic or constant), and the size of the in the upper half of the data. For example, $FC(1)$ is a structure with a constant mean and a constant variance of 1, whereas $SH(5)$ is a dataset with a stepwise-increasing mean and with a variance of $5^2$ for the upper half of the data. Sample datasets from $FH(3)$ and $SH(8)$ are shown in Figure 1 for illustration.
\begin{figure}
        \centering
        \begin{subfigure}[b]{0.45\textwidth}
                \centering
                \includegraphics[width=\textwidth]{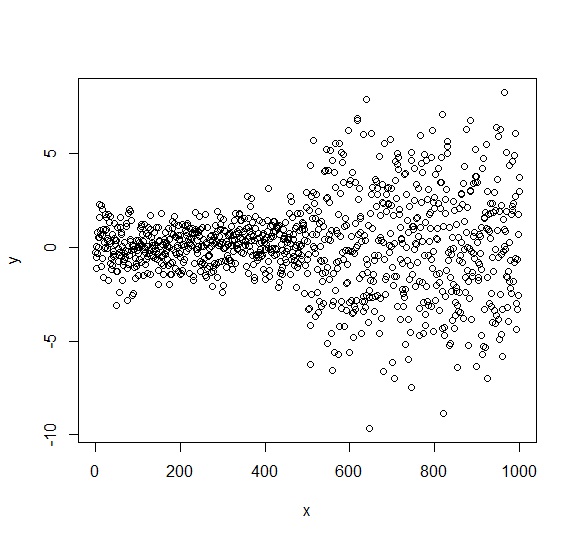}
                \caption{$FH(3)$}
        \end{subfigure}%
        ~
        \begin{subfigure}[b]{0.45\textwidth}
                \centering
                \includegraphics[width=\textwidth]{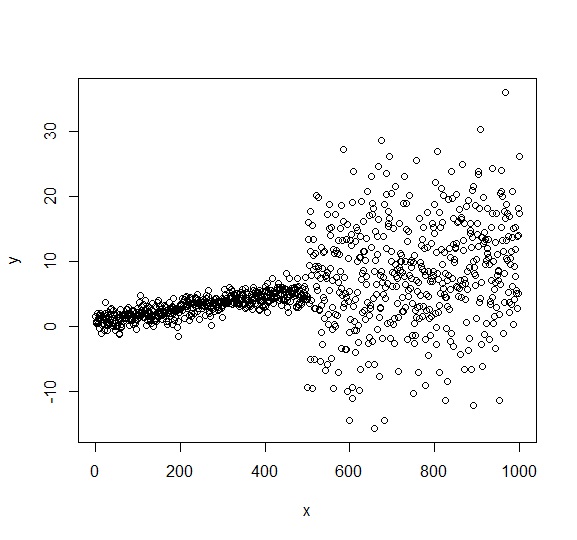}
                \caption{$SH(8)$}
        \end{subfigure}
        ~
        \caption{Sample Datasets}
\end{figure}
\subsection{Data Generation and Tree Fitting}
\label{dgtf}
All computations are performed in \texttt{R} \cite{r14} and make extensive use of the \texttt{rpart} package \cite{the14}. For each mean-variance combination, 10000 datasets are generated as follows.

We begin by fixing $x=1,...,1000$ and generate $Y_x$ to be independent normal random variables with mean $0$ and variance $1$. This gives us a sample dataset from the $FC(1)$ structure, where $Y_x = e_x$.  Call these the ``baseline errors" $e_x^{(0)}$. To obtain a sample dataset from each other structure, we multiply $e_1^{(0)},...e^{(0)}_{500}$ by $c_1$ and $e^{(0)}_{501},...,e^{(0)}_{1000}$ by $c_2$ then add $\mu_x$ to the results for $x=1,...,1000$. This procedure is then repeated for $j=1,...,10000$ to obtain 10000 sets of baseline errors and subsequent responses from each mean-variance structure. This is done instead of generating 10000 new datasets for each structure so that differences in split performance among different variance structures can be attributed solely to the changes in the variance, and not due to randomness in the data generation process.

We use the RPAB to construct a pruned regression tree fitting $Y$ by $X$ using the \texttt{rpart} function with default parameters (minimum SSE reduction of 0.01 and a minimum terminal node size of 7). On each tree, the performance measures listed below are computed.  These are then summarized across the 10,000 data sets for each mean-variance structure.

\subsection{Performance Measures}
\label{pm}

The ultimate goal of the regression is to accurately estimate the mean response.  A tree's ability to achieve this goal depends on the locations of its splits.   We are therefore interested in two main aspects of a tree's performance. We first look at a tree's ability to place splits near where they belong, and inversely, not to place splits where they don't belong. We then consider the tree's mean squared error (MSE) for estimating the true means under which its data are generated. 

In order to understand where the RPAB chooses to make splits, we look at the average number of splits made, and the distribution of where these splits occur under each framework. The average number of splits per dataset allows us to see how often spurious splits are made in the flat-mean case, and how many genuine splits are missed in the stepwise-increasing-mean case. In order to make this comparison, we compare heteroscedastic datasets to homoscedastic datasets with variance $(1+c_2^2)/2$ for reasons that we discuss in the next paragraph. The distribution of split locations allows us to find any patterns in how splits are either inappropriately made or omitted.

To examine the effect of heteroscedasticity on the MSE, we use the average total MSE under each framework, and the average MSE separately in each half of the dataset. The average total MSE gives us a ``global'' estimate of how well the RPAB is performing. We therefore compare heteroscedastic datasets to homoscedastic versions that have the same total variance across the entire range of $x$.  This requires setting the constant variance to $(1+c_2^2)/2$, which we refer to as the ``compromise'' variance. This structure has the same global variance as a heteroscedastic dataset with second-half standard deviation $c_2$, so we also use it to study the averge number of splits per dataset.

We also consider MSE comparisons where the ``local'' variance (i.e., the variance within the segment) is the same for the hetero- and homoscedastic cases in each segment.  This means that for each $c_2^2$, the average MSE in the lower five segments resulting from a model fit to heteroscedastic data are compared to the MSE in the lower half from trees fit to data sets with a constant variance of 1. Similarly, for each $c_2$, the average MSE in the upper five segments of a heteroscedastic dataset is compared to the average MSE from the fits to data with constant variance $c_2$. Thus, if the tree makes the same splits in both cases, the average MSEs for a given $c_2$ should be equal. This is because exactly the same means will be fit to the common data sets.

\subsection{Limitations}
\label{lim}

In this study, we investigate the performance of the RPAB on ideal, ``laboratory'' data. The purpose of this model is to allow us to identify where and why the disturbances might occur. While it is hardly a realistic model, simulating data where the variance jumps in the middle of the dataset allows us to clearly see how high variance data affects algorithm performance on the low variance data and vice-versa. Similarly, the stepwise-increasing-mean framework allows us to identify exactly where splits should be made. These results will not apply directly to ``real-world'' analyses, but instead serve as a warning of potential issues that may arise when the RPAB is applied blindly to heteroscedastic data.

\section{Results}

\subsection{Splits}
\label{spl}

The total numbers of splits over all datasets of a particular structure are compared in Table 1. These are presented as an average number of splits per dataset to make comparisons easier. For the constant mean case, the average number of splits increases rapidly as $c_2$ increases (recall that $c_2=1$ represents constant variance), and reaches a plateau of more than three times the number of splits made on a homoscedastic dataset (recall that the RPAB is scale invariant, so it will make the same splits on all homoscedastic flat-mean datasets).  However, the numbers are still very low. On the other hand, for the stepwise-increasing-mean structure, the average number of splits decreases as $c_2$ increases for both homoscedastic and heteroscedastic datasets.  The numbers do not differ drastically between the two cases.

\begin{table}[htbp]
  \centering
  \caption{Average number of splits per dataset}
    \resizebox{\textwidth}{!}{
    \begin{tabular}{rrrrrrrrrrr}
    \toprule
    Structure            & $c_2=1$ & $c_2=2$ & $c_2=3$ & $c_2=4$ & $c_2=5$ & $c_2=6$ & $c_2=7$ & $c_2=8$ &$c_2=9$ & $c_2=10$ \\
    \midrule
    $FH(c)$               & 0.02 & 0.06 & 0.07 & 0.07 & 0.07 & 0.07 & 0.07 & 0.07 & 0.08 & 0.08 \\
    $SC(\sqrt{(1+c^2)/2})$ & 5.0 & 4.7 & 4.0 & 3.5 & 3.3 & 3.1 & 2.9 & 2.7 & 2.5 & 2.3 \\
    $SH(c)$               & 5.0 & 4.7 & 4.1 & 3.4 & 3.1 & 3.0 & 2.9 & 2.8 & 2.7 & 2.5 \\
    \bottomrule
    \end{tabular}%
    }
\end{table}%

Figures 2, 3 and 4 show plots of the total number of splits at each value of $X$ across the $10,000$ datasets for structures $FC(c)$, $SC(c)$ and $SH(c)$. The values $c=1,5,10$ are chosen to give a reasonable representation of the main trends.

In the constant-mean structure (Figure 2), the effect of increasing variance in the right half of the data is clear.  Under homoscedasticity, the plot is fairly symmetric with a tendency for more splits to occur near either extreme (this ``end-cut preference'' has been documented in Breiman et al.~\cite{bre84}). When the variance in the upper half of the data is much larger than in the lower half, no splits occur in the region of low variability, and the split frequency in the region of higher variability is increased substantially.

For homoscedastic data with a stepwise-increasing-mean, low-variance datasets tend to split almost exclusively at or around each mean jump (Figure 3). On average, about half of the jumps are detected when the standard deviation is equal to the step size (Table 1). As the variance increases uniformly across $x$, splits at the extreme jumps are lost and those around other jumps become more diffuse.  Interestingly, splits at 400 and 600 are also suppressed.  Eventually, when the variance difference is extremely large, fewer splits are made anywhere. There is a clear preference for splits near the center of the range, but splits are no longer very focused near the actual jumps; rather, they occur more uniformly across entire segments.

We see a similar pattern under the heteroscedastic stepwise-increasing framework. One obvious difference is that splits around 100 and 400 become less frequent as the severity of the heteroscedasticity increases and eventually splits around 100 are missed entirely.
\begin{figure}
\centering
\includegraphics[height=0.3\paperheight]{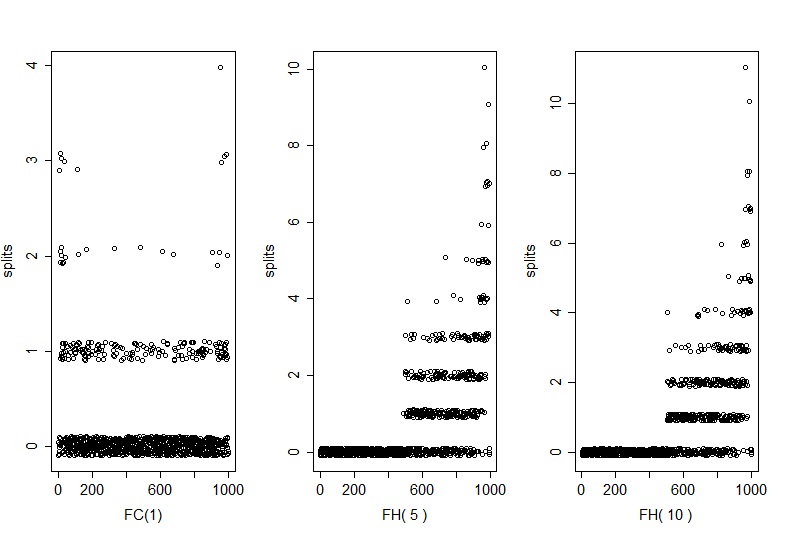}
\caption{Plots of the number of splits at $X=x$ against $x$ under the $FH(c)$ framework for $c=1,5,10$. Note that the scales of the plots are left different to highlight patterns.}
\end{figure}

\begin{figure}
\centering
\includegraphics[height=0.3\paperheight]{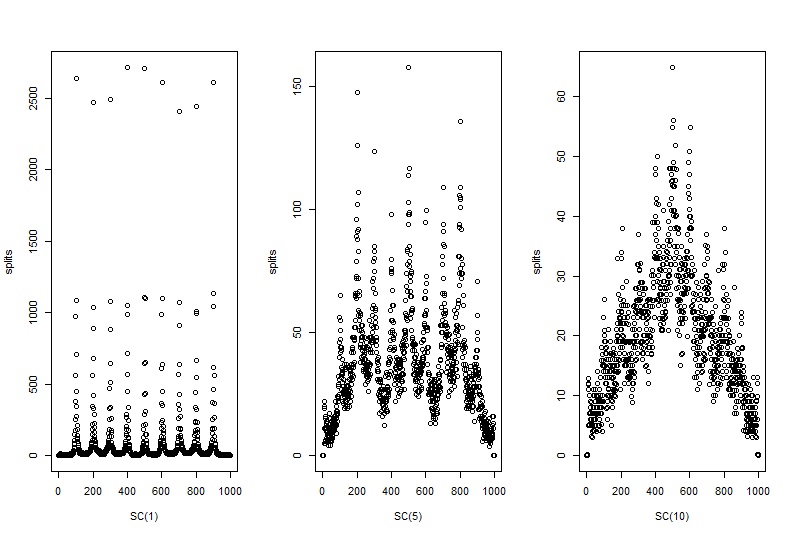}
\caption{Plots of the number of splits at $X=x$ against $x$ under the $SC(c)$ framework for $c=1,5,10$. Note that the scales of the plots are left different to highlight patterns.}
\end{figure}

\begin{figure}
\centering
\includegraphics[height=0.3\paperheight]{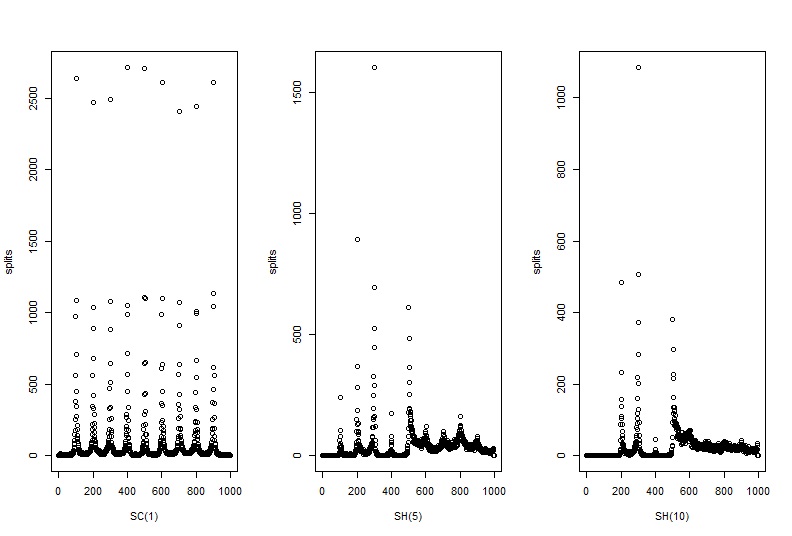}
\caption{Plots of the number of splits at $X=x$ against $x$ under the $SH(c)$ framework for $c=1,5,10$. Note that the scales of the plots are left different to highlight patterns.}
\end{figure}

Figure 5 gives a more detailed view of the number of splits for the second half of the $SC(10)$ and $SH(10)$ plots. Note that many more splits occur near $500$ in the heteroscedastic case. This suggests a small stabilizing effect of the low variance half. We will discuss this more later.

\begin{figure}
\centering
\includegraphics[width=0.7\textwidth]{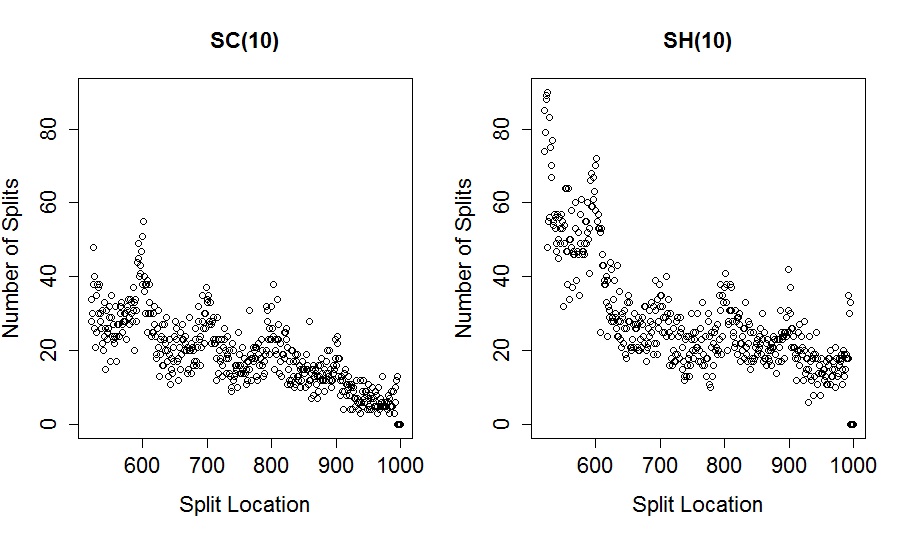}
\caption{Comparison of split locations between the second halves for $SC(10)$ and $SH(10)$.}
\end{figure}

\subsection{Mean Squared Errors}
\label{mse}

We compute the average MSE per dataset under the heteroscedastic and compromise variance structures for $c_2=1,...,10$. The ratios of the averages for heteroscedastic datasets to the averages for compromise datasets are presented in Table 2 for both mean structures. Here we see that heteroscedasticity has a stronger effect for larger values of $c_2$. Further simulations (not shown) indicate that this trend continues for larger values of $c_2$. Further, the effect is more pronounced for homoscedastic datasets.

\begin{table}[htbp]
  \centering
  \caption{Ratio of average total MSEs between heteroscedastic and compromise variance structures.}
    \resizebox{\textwidth}{!}{
    \begin{tabular}{rrrrrrrrrrr}
    \toprule
    Structure & $c_2=1$ & $c_2=2$ & $c_2=3$ & $c_2=4$ & $c_2=5$ & $c_2=6$ & $c_2=7$ & $c_2=8$ &$c_2=9$ & $c_2=10$ \\
    \midrule
    Flat-Mean & 1.00 & 1.64 & 1.86 & 1.99 & 1.99 & 2.02 & 1.98 & 1.99 & 1.99 & 1.97  \\
    Stepwise-Mean & 1.00 & 0.97 & 0.96 & 1.06 & 1.10 & 1.09 & 1.08 & 1.08 & 1.13 & 1.25 \\
    \bottomrule
    \end{tabular}
    }
\end{table}


\begin{figure}
\centering
\begin{subfigure}[t]{0.45\textwidth}
\includegraphics[width=\textwidth]{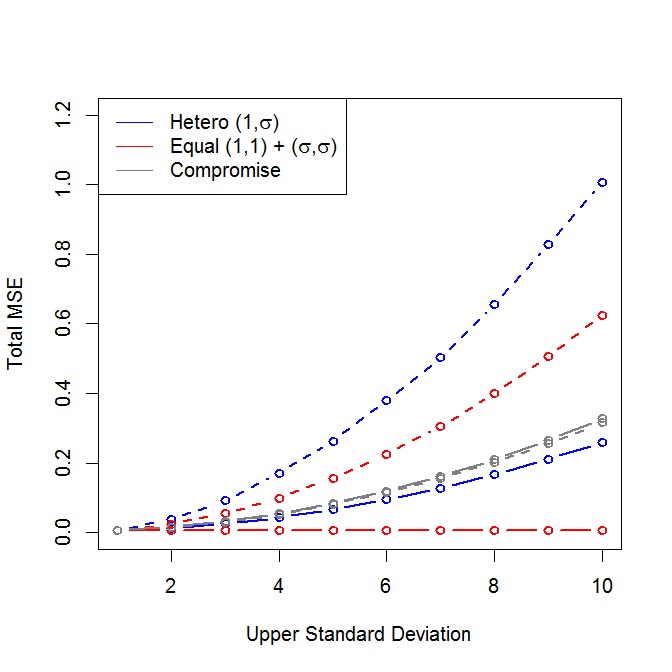}
\caption{Flat-mean.}
\end{subfigure}
\hfill
\begin{subfigure}[t]{0.45\textwidth}
\includegraphics[width=\textwidth]{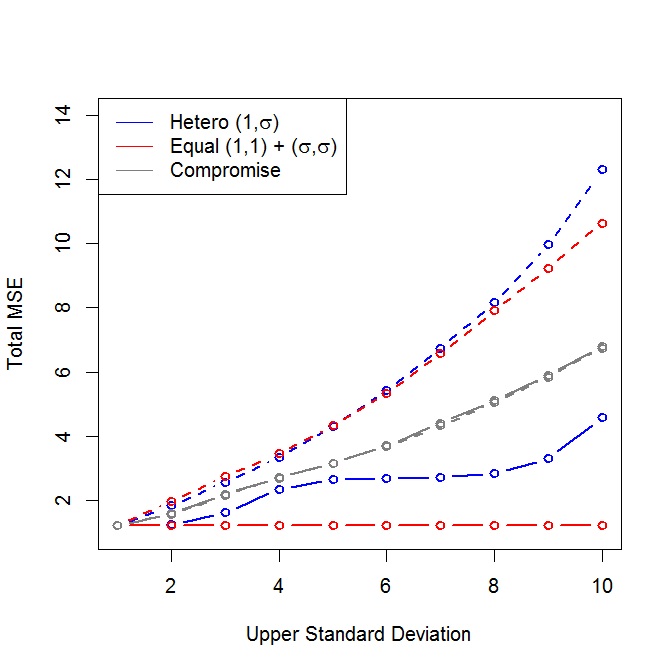}
\caption{Stepwise-mean.}
\end{subfigure}
\caption{Total MSEs by dataset half for each mean and variance structure as second half standard deviation increases. Solid and dashed lines correspond to the first and second halves respectively.}
\end{figure}

In Figure 6(a), we see the expected result that all MSEs increase as the variance increases, except for the first-half homoscedastic case, where the variance is constantly 1.  Notice that the MSEs are consistently higher in the heteroscedastic case than in their homoscedastic counterparts.  This is a direct measure of the effect of the extra splits that are made in the heteroscedastic cases; it is interesting that they impact both halves of the data. Of course,  the compromise MSEs are approximately identical between the two halves.

In Figure 6(b) we see a somewhat different pattern for stepwise-increasing datasets. The total MSE in the second half of the datasets is more similar for heteroscedastic and homoscedastic structures. This suggests that the loss of precision in the split locations has an overall similar effect in the two variance structures, reflecting the similarity of patterns in Figure 5. The lower variance in the first half does not provide a substantial stabilizing effect on the second. However, the higher variance does destabilize estimation in the low-variance half, as seen in the two curves for the lower-half MSEs. This is directly traceable to the loss of splits at actual jumps in the lower half of the data. Again, we see that the two compromise lines are almost identical. Notice that the sum of the two lines at any variance level is similar to or less than the sum for the two halves from the heterscedastic case.  This corroborates the trends seen in Table 2.

The MSE of the first half in the heteroscedastic case is the least regular of the six curves. The two rapid increases joined by a fairly stable region is due to the RPAB choosing to make fewer splits in the first half as the variance in the second half increases. As the total variance increases, the relative improvement in the MSE provided by a split in the first half decreases. The regions with rapid increases in total MSE are therefore the thresholds at which the RPAB reduces the number of splits it makes in the first half.

\section{How Splits Affect the MSE}
\label{hsam}

Under the mean-variance structures we consider, it is relatively easy to derive exact MSE results for certain simple situations, like considering the effect of a single split. For example, it can be shown that, for trees fit to heteroscedastic data, spurious splits can actually improve estimation in some regions of the regression; provided that the larger variance is sufficiently large that the standard error of the mean from all observations is larger than that from just the ones with smaller variance (here, this happens when $c_2^2 > 3c_1^2$). However, the drawback is that in other regions of the regression, spurious splits make estimation {\em less} precise, so that the average MSE across the entire fit is higher.

This argument can be extended to stepwise-increasing-mean structures to explain why some ``obvious'' splits are missed, particularly in low-variance regions of heteroscedastic data. When the data are split into two groups, then the mean of each group is estimated with fewer points. This results in a larger variance contribution to the MSE. This increase is balanced by a reduction in bias from ensuring that means are estimated by points whose true means are not too disparate. This trade-off leads to certain splits not being made if the decrease in bias is not large enough to offset the increase in variance. This is particularly apparent if the change in mean is small relative to the local variance.

\section{Discussion and Conclusion}
\label{dc}

The average number of splits per dataset with the constant mean structure increases quite rapidly then levels off as the second half standard deviation increases. The most obvious pattern that arises in the locations of these splits is the preference for splits to occur in the second half of a dataset as the second half variance gets larger. To see why this occurs, first note that the amount of improvement in the SSE criterion that occurs with a split in the first half of the data does not change as the variance of the second half increases. However, the total SSE of the dataset does increase with the variance of the second half. This means that the proportion of the total SSE that can be explained by a split in the first half decreases with $c_2$. Meanwhile, much larger changes in the SSE criterion are often available by collecting a small number of observations in the high-variance region into a common mean that is quite different from that of the other points in the region.  Due to the nature of the splitting algorithm, this is easier to do with points near the edge of the regression space.  This latter phenomenon is discussed in Breiman et al.~\cite{bre84}.

What happens to splits in low-variance regions of the stepwise-increasing-mean structure when variance in distant regions grows is a major concern.  Splits that would seem obvious to the naked eye are not made by the RPAB.  To explore this further, we reduced the variance to zero in the low-variance half and re-analyzed the data for various values of $c_2$ (details not shown).  Plots of the split locations still showed a failure to detect these obvious jumps, with a similar general pattern to Figure 4!  We wondered whether the default settings in \texttt{rpart} were interfering with split selections by terminating the algorithm too early, because the large variance in the upper half could create a total SSE so large that no split in the lower half could possibly result in a 1\% decrease in the criterion.  We therefore reduced the minimum improvement needed to split, while maintaining a minimum terminal node size of 7, so that all jumps would be detected nearly perfectly. We then pruned the tree as usual.  We discovered that the cost-complexity pruning algorithm removed nearly all of the obvious splits.  This points to an immediate need for a better pruning algorithm to augment the current splitting algorithm, as well as the need to manually adjust the default settings in this popular software so that proper splits are detected and retained.

Under the stepwise-increasing-mean structure, the average number of splits per dataset decreases with the second half standard deviation in both homoscedastic and heteroscedastic datasets. This indicates declining performance because the algorithm should be splitting at each of the nine mean jumps. The locations of these splits also become less reasonable as the second half standard deviation increases.

In Section \ref{spl} we noted the strange pattern of preference for splits at 200 or 300 and reluctance to split at 100 or 400 in heteroscedastic step-wise mean cases. This occurs because the algorithm tends to split at or near 500 first.  This location maximally reduces the bias in the two resulting nodes while having the smallest effect on the estimation variance. After that, the most bias-reducing split in the lower half of the data again occurs near the middle.  For our data structure, the optimal places to split are at the two jumps near the middle of the lower half of the data; i.e. at 200 or 300 with roughly equal probability.  Successive splits can be made within each subsequent node, with jump points being optimal locations.  However, the improvements in the SSE criterion at each successive split are incrementally smaller than what is gained from the initial split.  In data with much larger variance elsewhere, potentially bigger gains are available by making splits in the high-variance portion of the data,  even when those splits may not be necessary.  This may explain why the pruning algorithm does not retain obvious splits in low-variance regions: they provide less reduction in the SSE criterion than splits in higher-variance region whose utility is questionable.  If the latter splits are pruned away, then the former splits are as well.

In conclusion, the results of our analyses clearly indicate that the RPAB can be sensitive to heteroscedasticity.  Its OLS-based pruning criterion is particularly problematic and can cause splits that are obvious to the eye to be lost in the final tree.  The consequences of this sensitivity may or may not extend to ensemble methods that use the RPAB to identify the individual members of the ensemble, such as random forests and boosted trees.  These approaches typically don't use pruning, which is computationally much more expensive than splitting.  However, boosting in particular tends to grow fairly small trees at each iteration, and thus might be expected to consistently miss obvious splits that contribute relatively small improvements to the overall criterion in favor of possibly spurious splits in regions of high variance.  Further study is needed to assess and to correct this potential loss.

\section*{Acknowledgements}

This research was supported by the National Science and Engineering Research Council of Canada (NSERC).

\bibliographystyle{plain}

\bibliography{references}






\end{document}